\documentclass[11pt,letterpaper]{article}
\usepackage[letterpaper]{geometry}
\usepackage{acl2012}
\usepackage{amsmath}
\usepackage{booktabs}
\usepackage{color}
\usepackage{latexsym}
\usepackage{microtype}
\usepackage{multirow}
\usepackage{times}
\usepackage{url}
\usepackage{comment}
\usepackage[pdftex]{graphicx}
\usepackage{cleveref}

\makeatletter
\newcommand{\@BIBLABEL}{\@emptybiblabel}
\newcommand{\@emptybiblabel}[1]{}
\makeatother
\usepackage[hidelinks]{hyperref}

\newcommand{\cell}[1]{\begin{tabular}[t]{@{}l@{}}#1\end{tabular}}

\title{The NarrativeQA Reading Comprehension Challenge}

\author{
Tom\'a\v s Ko\v cisk\'y$^{\dag\ddag}$ \quad
Jonathan Schwarz$^{\dag}$ \quad 
Phil Blunsom$^{\dag\ddag}$ \quad 
Chris Dyer$^{\dag}$ \\
{\bf Karl Moritz Hermann$^{\dag}$ \quad 
G\'abor Melis$^{\dag}$ \quad 
Edward Grefenstette$^{\dag}$} \\
$^{\dag}$DeepMind \quad
$^{\ddag}$University of Oxford \\
{\tt \{tkocisky,schwarzjn,pblunsom,cdyer,kmh,melisgl,etg\}@google.com}
}

\date{}

\newcommand{\examplea}[7]{%
\begin{figure}[tb]
\begin{tabular}{p{0.92\columnwidth}}
\toprule
\noindent%
\textbf{Title:} #1\\
\textbf{Question:} #2\\
\textbf{Answer:} #3\\
\textbf{Summary snippet:} #4\\
\textbf{Story snippet:} #5 \\
\bottomrule
\end{tabular}
\caption{#6}
\label{#7}
\end{figure}%
}

\newcommand{\ghostexample}{%
\examplea{Ghostbusters II}
{\small How is Oscar related to Dana?}
{\small her son}
{\small\dots Peter's former girlfriend Dana Barrett has had a son, Oscar\dots}
{\\
\small \centerline{\textit{DANA (setting the wheel brakes on the buggy)}}
\small \centerline{Thank you, Frank.  I'll get the hang of this eventually.}

\vspace{0.2cm}
\small She continues digging in her purse while Frank leans over the buggy and makes funny faces at the baby, OSCAR, a very cute nine-month old boy.

\vspace{0.2cm}
\small\centerline{\textit{FRANK (to the baby)}}
\small\centerline{Hiya, Oscar.  What do you say, slugger?}

\vspace{0.2cm}
\small\centerline{\textit{FRANK (to Dana)}}
\small That's a good-looking kid you got there, Ms.\ Barrett.}
{Example question--answer pair. The snippets here were extracted by humans from summaries and the full text of movie scripts or books, respectively, and are \emph{not} provided to the model as supervision or at test time. Instead, the model will need to read the full text and locate salient snippets based solely on the question and its reading of the document in order to generate the answer.}
{fig:example1}}

\newcommand{\exasizea}{\small}

\newcommand{\armageddonexample}{%
\examplea{Armageddon 2419 A.D.}
{\exasizea In what year did Rogers awaken from his deep slumber?}
{\exasizea 2419}
{\exasizea\dots Rogers remained in sleep for 492 years. He awakes in 2419 and,\dots}
{\exasizea I should state therefore, that I, Anthony Rogers, am, so far as I know,
the only man alive whose normal span of eighty-one years of life has
been spread over a period of 573 years. To be precise, I lived the first
twenty-nine years of my life between 1898 and 1927; the other fifty-two
since 2419. The gap between these two, a period of nearly five hundred
years, I spent in a state of suspended animation, free from the ravages
of katabolic processes, and without any apparent effect on my physical
or mental faculties. When I began my long sleep, man had just begun his real conquest of the air\dots}
{Example question--answer pair with snippets from the summary and the story.}
{fig:example2}%
}

\newcommand{\jacobexample}{%
\examplea{Jacob's Ladder}
{\exasizea What is the fatal injury that Jacob sustains which ultimately leads to his death ?}
{\exasizea A bayonete stabbing to his gut.}
{\exasizea A terrified Jacob flees into the jungle, only to be bayoneted in the gut by an unseen assailant.

    \vspace{-0.2cm}
    \centerline{$[\dots]$}
    
    In a wartime triage tent in 1971, military doctors fruitlessly treating Jacob reluctantly declare him dead}
{\exasizea As he spins around 
   one of the attackers jams all eight inches of his bayonet blade into 
   Jacob's stomach. Jacob screams. It is a loud and piercing wail.
    
    \vspace{-0.2cm}
    \centerline{$[\dots]$}
    
    \exasizea \centerline{\textit{Int. Vietnam Field Hospital - Day}}
    
    A doctor leans his head in front of the lamp and removes his 
    mask. His expression is somber. He shakes his head. His words are 
    simple and final.
    
    \medskip
    \exasizea \centerline{\textit{DOCTOR}}
    		\centerline{He's gone.}
    
    \medskip		
    \exasizea \centerline{\textit{Cut to Jacob Singer} \dots}
    
    The doctor steps away. A nurse rudely 
    pulls a green sheet up over his head. The doctor turns to one of 
    the aides and throws up his hands in defeat.}
{Example question--answer pair with snippets from the summary and the story.}
{fig:example3}
}

\begin{document}
\maketitle
\begin{abstract}
Reading comprehension (RC)---in contrast to information retrieval---requires integrating information and reasoning about events, entities, and their relations across a full document. Question answering is conventionally used to assess RC ability, in both artificial agents and children learning to read. However,
existing RC datasets and tasks are dominated by questions that can be solved by selecting answers using superficial information (e.g., local context similarity or global term frequency); they thus fail to test for the essential integrative aspect of RC. To encourage progress on deeper comprehension of language, we present a new dataset and set of tasks in which the reader must answer questions about stories by reading entire books or movie scripts. These tasks are designed so that successfully answering their questions requires understanding the underlying narrative rather than relying on shallow pattern matching or salience. We show that although humans solve the tasks easily, standard RC models struggle on the tasks presented here. We provide an analysis of the dataset and the challenges it presents.
\end{abstract}

\section{Introduction}

\ghostexample

Natural language understanding seeks to create models that read and comprehend text. A common strategy for assessing the language understanding capabilities of comprehension models is to demonstrate that they can answer questions about documents they read, akin to how reading comprehension is tested in children when they are learning to read.
After reading a document, a reader usually can not reproduce the entire text from memory, but often can answer questions about underlying narrative elements of the document: the salient entities, events, places, and the relations between them.
Thus, testing understanding requires creation of questions that examine high-level abstractions instead of just facts occurring in one sentence at a time.

Unfortunately, superficial questions about a document may often be answered successfully (by both humans and machines) using a shallow pattern matching strategies or guessing based on global salience. 
In the following section, we survey existing QA datasets, showing that they are either too small or answerable by shallow heuristics~(Section~\ref{sec:review}). On the other hand, questions which are not about the surface form of the text, but rather about the underlying narrative, require the formation of more abstract representations about the events and relations expressed in the course of the document. Answering such questions requires that readers integrate information which may be distributed across several statements throughout the document, and generate a cogent answer on the basis of this integrated information. That is, they test that the reader comprehends language, not just that it can pattern match. We present a new task and dataset, which we call NarrativeQA, which will test and reward artificial agents approaching this level of competence (Section~\ref{sec:narrativeqa}).

The dataset consists of \emph{stories}, which are books and movie scripts, with human written questions and answers based solely on human-generated abstractive \emph{summaries}. For the RC tasks, questions may be answered using just the summaries or the full story text. We give a short example of a sample movie script from this dataset in Figure~\ref{fig:example1}.
Fictional stories have a number of advantages as a domain. First, they are largely self-contained: beyond the basic fundamental vocabulary of English, all the information about salient entities and concepts required to understand the narrative is present in the document, with the expectation that a reasonably competent language user would be able to understand it.\footnote{For example, new names and words may be coined by the author (e.g.~``muggle'' in Harry Potter novels) but the reader need only appeal to the book itself to understand the meaning of these concepts, and their place in the narrative. This ability to form new concepts based on the contexts of a text is a crucial aspect of reading comprehension, and is in part tested as part of the question answering tasks we present.} Second, story summaries are abstractive and generally written by independent authors who know the work only as a reader.
We make the dataset available online.\footnote{\url{http://deepmind.com/publications}}

\section{Review of Reading Comprehension Datasets and Models}
\label{sec:review}

\begin{table*}[tbh]
\scriptsize
\centering
\begin{tabular}{p{5.1cm}p{4.0cm}p{3.2cm}p{2.5cm}}
    \toprule
    \textbf{Dataset} & \textbf{Documents} & \textbf{Questions} & \textbf{Answers}  \\
    \midrule
    MCTest \cite{mctest} &
        \cell{660 short stories,\\grade school level} &
        \cell{2640 human generated,\\based on the document} &
        multiple choice \\
    CNN/Daily Mail \cite{nips15hermann} &
        93K+220K news articles &
        \cell{387K+997K Cloze-form,\\based on highlights} &
        entities \\
    Children's Book Test (CBT) \cite{hill-cbt} &
        687K of 20 sentence passages from 108 children's books &
        \cell{Cloze-form,\\from the 21st sentence} &
        multiple choice \\
    BookTest \cite{booktest} &
        14.2M, similar to CBT &
        Cloze-form, similar to CBT &
        multiple choice \\
    SQuAD \cite{squad} &
        23K paragraphs from 536 Wikipedia articles &
        \cell{108K human generated,\\based on the paragraphs}&
        spans \\
    NewsQA \cite{maluuba-NewsQA} &
        13K news articles from the CNN dataset &
        \cell{120K human generated,\\based on headline, highlights}&
        spans\\
    MS MARCO \cite{msmarco} &
        1M passages from 200K+ documents retrieved using the queries&
        100K search queries&
        \cell{human generated,\\ based on the passages}\\
    SearchQA \cite{dunn2017searchqa} &
        6.9m passages retrieved from a search engine using the queries&
        140k human generated \mbox{Jeopardy!} questions &
        human generated Jeopardy! answers\\
    \midrule
    NarrativeQA (this paper) &
        1,572 stories (books, movie scripts) \& human generated summaries &
        \cell{46,765 human generated, \\based on summaries}&
        \cell{human generated,\\ based on summaries} \\
    \bottomrule
\end{tabular}
\caption{Comparison of datasets.}
\label{tab:related_datasets}
\end{table*}

There are a large number of datasets and associated tasks available for the training and evaluation of reading comprehension models. We summarize the key features
of a collection of popular recent datasets in Table~\ref{tab:related_datasets}. In this section, we briefly discuss the nature and limitations of these datasets and their associated tasks.

MCTest~\cite{mctest} is a collection of short stories, each with multiple questions. Each such question has set of possible answers, one of which is labelled as correct.
While this could be used as a QA task, the MCTest corpus is in fact intended as an answer selection corpus. The data is human generated, and the answers can be phrases or sentences. The main limitation of this dataset is that it serves more as a an evaluation challenge than as the basis for end-to-end training of models, due to its relatively small size.

In contrast, CNN/Daily Mail~\cite{nips15hermann}, Children's Book Test (CBT)~\cite{hill-cbt}, and BookTest \cite{booktest} each provide large amounts of question--answer pairs. Questions are Cloze-form (predict the missing word) and are produced from either short abstractive summaries (CNN/Daily Mail) or from next sentence in the document the context was taken from (CBT and BookTest). The tasks associated with these datasets are all selecting an answer from a set of options, which is explicitly provided for CBT and BookTest, and is implicit for CNN/Daily Mail, as the answers are always entities from the document.
This significantly favors models that operate by pointing to a particular token (or type). Indeed, the most successful models on these datasets, such as the Attention Sum Reader (AS Reader)~\cite{kadlec2016text}, exploit precisely this bias in the data. However, these models are inappropriate for answers requiring synthesis of a new answer. This bias towards answers that are shallowly salient is a more serious limitation of the CNN/Daily Mail dataset, since its context documents are news stories which usually contain a small number of salient entities and focus on a single event.

SQuAD~\cite{squad} and NewsQA \cite{maluuba-NewsQA} offer a different challenge. A large number of a questions and answers are provided for a set of documents, where the answers are \emph{spans} of the context document, i.e.~contiguous sequences of words from the document. Although the answers are not just single word/entity answers, many plausible questions for assessing RC cannot be asked because no document span would contain its answer.
While they provide a large number of questions, these are from a relatively small number of documents, which are themselves fairly short, thereby limiting the lexical and topical diversity models trained on this data can cope with. While the answers are multi-word phrases, the spans are generally short
and rarely cross sentence boundaries.
Simple models scoring and/or extracting candidate spans conditioned on the question and superficial signal from the rest of the document do well
\cite[e.g.]{seo2016bidirectional}.
These models will not trivially generalize to problems where the answers are not spans in the document, supervision for spans is not provided, or several discontinuous spans are needed to generate a correct answer.
This restricts the scalability and applicability of models doing well on SQuAD or NewsQA to more complex problems.

MS MARCO~\cite{msmarco} presents a bolder challenge: questions are paired with sets of snippets (``context passages'') that contain the information necessary to answer the question,
and answers are free-form human generated text. However, as no restriction was placed on annotators preventing them from copying answers from source documents, many answers are in fact verbatim copies of short spans from the context passages.
Models which do well on SQuAD (e.g.\ \newcite{wang2016machine}, \newcite{Weissenborn2017FastQAAS}), extracting spans or pointing, do well here too, and the same concerns as above about the general applicability of solutions to this dataset to larger reading comprehension problems applies.

SearchQA \cite{dunn2017searchqa} is a recent dataset in which the context for each question is a set of documents retrieved by a search engine using the question as the query. However, in contrast with previous datasets neither questions nor answers were produced by annotating the context documents, but rather the context documents were retrieved after collecting pre-existing question--answer pairs.
As such, it is not open to same annotation bias as the datasets discussed above. However, upon examining answers in the Jeopardy data used to construct this dataset, one finds that 80\% of answers are bigrams or unigrams, and 99\% are 5 tokens or fewer. Of a sample of 100 answers, 72\% are named entities, all are short noun-phrases.

\paragraph{Summary of Limitations.}
We see several limitations of the scope and depth of the RC problems in existing datasets. First, several datasets are small (MCTest) or not overly naturalistic (bAbI; \newcite{babi}).
Second, in more naturalistic documents, a majority of questions require only a single sentence to locate supporting information for answering \cite{chen_examine_cnn,squad}. This, we suspect, is largely an artifact of the question generation methodology, in which annotators have created questions from a context document, or where context documents that explicitly answer a question are identified using a search engine. Although the factoid-like Jeopardy questions of SearchQA also appears to favor questions answerable with local context.
Finally, we see further evidence of the superficiality of the questions in the architectures that have evolved to solve them, which tend to exploit span selection based on representations derived from local context and the query \cite{seo2016bidirectional,rnet}.

\section{NarrativeQA: A New Dataset}
\label{sec:narrativeqa}

In this section, we introduce our new dataset, NarrativeQA, which addresses many of the limitations identified in existing datasets.
\subsection{Desiderata}
\label{sec:desiderata}

From the above discussed features and limitations, we define our desiderata as follows. We wish to construct a dataset with a large number of question--answer pairs based on either a large number of supporting documents or from a smaller collection of large documents. This permits the training of neural network-based models over word embeddings and provide decent lexical coverage and diversity. The questions and answers should be natural, unconstrained, and human generated, and answering questions should frequently require reference to several parts or a larger span of the context document rather than superficial representations of local context. Furthermore, we want annotators to privilege writing answers expressed in their own words, and consider higher-level relations between entities, places, and events, rather than copy short spans of the document.

Furthermore, we want to evaluate models both on the fluency and correctness of generated free-form answers, and as an answer selection problem, which requires the provision of sensible distractors to the correct answer. Finally, the scope and complexity of the QA problem should be such that current models struggle, while humans are capable of solving the task correctly, so as to motivate further research into the development of models seeking human reading comprehension ability.

\subsection{Data Collection Method}

We will consider complex, self-contained narratives as our documents/stories. To make the annotation tractable and lead annotators towards asking non-localized questions, we will only provide them human written summaries of the stories for generating the question--answer pairs.

We present both books and movie scripts as stories in our dataset. Books were collected from Project Gutenberg\footnote{\url{http://www.gutenberg.org/}} and movie scripts scraped from the web.\footnote{Mainly from \url{http://www.imsdb.com/}, but also \url{http://www.dailyscript.com/}, \url{http://www.awesomefilm.com/}.}
We matched our stories with plot summaries from Wikipedia using titles and verified the matching with help from human annotators. The annotators were asked to determine if both the story and the summary refer to a movie or a book (as some books are made into movies), or if they are the same part in a series produced in the same year. In this way we obtained 1,567 stories. This provides with a smaller set of documents, compared to the other datasets, but the documents are long which provides us with good lexical coverage and diversity. The bottleneck for obtaining a larger number of publicly available stories was finding corresponding summaries.

Annotators on Amazon Mechanical Turk were instructed to write 10 question--answer pairs each based solely on a given summary.
Reading and annotating summaries is tractable unlike writing questions and answers based on the full stories, and moreover, as the annotators never see the full stories we are much less likely to get questions and answers which are extracted from a localized context.

Annotators were instructed to imagine that they are writing questions to test students who have read the full stories but not the summaries. We required questions that are specific enough, given the length and complexity of the narratives, and to provide a diverse set of questions about characters, events, why this happened, and so on. Annotators were encouraged to use their own words and we prevented them from copying.\footnote{This was done both through instructions and Javascript hard limitations on the annotation site.}
We asked for answers that are grammatical, complete sentences, and explicitly allowed short answers (one word, or a few-word phrase, or a short sentence) as we think that answering with a full sentence is frequently perceived as artificial when asking about factual information.
Annotators were asked to avoid extra, unnecessary information in the question or the answer, and to avoid yes/no questions or questions about the author or the actors.

About 30~question--answer pairs per summary were obtained. The result is a collection of human written natural questions and answers. As we have multiple questions per summary/story, this allows us to consider answer selection (from among the 30) as a simpler version of the QA than answer generation from scratch. Answer selection \cite{P16-1145} and multiple-choice question answering \cite{mctest,hill-cbt} are frequently used.

We additionally collected a second reference answer for each question by asking annotators to judge whether a question is answerable, given the summary, and provide an answer if it was. All but 2.3\% of the questions were judged as answerable.

\subsection{Core Statistics}

We collected 1,567 stories, evenly split between books and movie scripts.
We partitioned the dataset into non-overlapping training, validation, and test portions,
along stories/summaries. See Table~\ref{tab:datasetstats} for detailed statistics.

The dataset contains 46,765 question--answer pairs. The questions are grammatical questions written by human annotators, average~9.8 tokens in length, and are mostly formed as `WH'-questions (see Table~\ref{tab:first_token}). We categorized a sample of 300 questions in Table~\ref{tab:question_categories}. We observe a good variety of question types. An interesting category are questions which ask for something related to or occurring together/before/after with an event, of which there are about 15\%.

Answers in the dataset are human written, short, averaging 4.73 tokens, but not restricted to spans from the documents.
There are 44.05\% and 29.57\% answers that appear as spans of the summaries and the stories, respectively; as expected, lower proportion of answers are spans on stories compared to summaries on which they were constructed.

\begin{table*}[t]
\begin{minipage}[t]{.48\linewidth}
\small
\footnotesize
\centering
\newcommand{\tokens}{tok.}
\begin{tabular}[t]{@{}llll@{}}
    \toprule
                                &  \textbf{train}   &  \textbf{valid}   &   \textbf{test}    \\
    \midrule
    \# documents                & 1,102    & 115      & 355       \\
    \dots\ books                & 548      & 58       & 177      \\
    \dots\ movie scripts        & 554      & 57       & 178       \\
    \# question--answer pairs    & 32,747   & 3,461    & 10,557    \\
    Avg. \#\tokens\ in summaries & 659      & 638      & 654       \\
    Max  \#\tokens\ in summaries & 1,161     & 1,189   & 1,148    \\
    Avg. \#\tokens\ in stories   & 62,528   & 62,743   & 57,780   \\
    Max  \#\tokens\ in stories   & 430,061  & 418,265  & 404,641    \\
    Avg. \#\tokens\ in questions & 9.83     & 9.69     & 9.85       \\
    Avg. \#\tokens\ in answers   & 4.73     & 4.60     & 4.72       \\
    \bottomrule
\end{tabular}
\caption{NarrativeQA dataset statistics.} \label{tab:datasetstats}
\end{minipage}
\hfill
\begin{minipage}[t]{.25\linewidth}
\small
\footnotesize
\centering
\begin{tabular}[t]{@{}lr@{}}
    \toprule
    \textbf{First token} & \textbf{Frequency} \\
    \midrule
    What&	38.04\% \\
    Who&	23.37\% \\
    Why&	~9.78\% \\
    How&	~8.85\% \\
    Where&	~7.53\% \\
    Which&	~2.21\% \\
    How many/much& 1.80\% \\
    When&	~1.67\% \\
    In&	    ~1.19\% \\
    OTHER& ~5.57\% \\
    \bottomrule
\end{tabular}
\caption{Frequency of first token of the question in the training set.}
\label{tab:first_token}
\end{minipage}%
\hfill
\begin{minipage}[t]{.24\linewidth}
\small
\footnotesize
\centering
\begin{tabular}[t]{@{}lr@{}}
    \toprule
    \textbf{Category} & \textbf{Frequency} \\
    \midrule
Person	&30.54\%  \\
Description~~~	&24.50\%  \\
Location	&~9.73\%  \\
Why/reason	&~9.40\%  \\
How/method	&~8.05\%  \\
Event	&~4.36\%  \\
Entity	&~4.03\%  \\
Object	&~3.36\%  \\
Numeric	&~3.02\%  \\
Duration	&~1.68\%  \\
Relation	&~1.34\%  \\
\bottomrule
\end{tabular}
\caption{Question categories on a sample of 300 questions from the validation set.}
\label{tab:question_categories}
\end{minipage}
\end{table*}

\subsection{Tasks}

We present
tasks varying in their scope and complexity: we consider either the summary or the story as context, and for each we evaluate answer generation and answer selection.

The task of answering questions based on summaries
is similar  in  scope  to previous datasets. However, summaries contain more complex relationships and timelines
than news articles or short paragraphs from the web and thus provide a task different in nature.
We hope that NarrativeQA will motivate the design of
architectures capable of modeling such relationships.
This setting is similar to the previous tasks in
that the questions and answers were constructed based on these supporting documents.

The full version of NarrativeQA requires reading and understanding entire stories (i.e., books and movie scripts).
This task is at present intractable for existing neural models out of the box. We further discuss the challenges and possible approaches in the following sections.

We require the use of metrics for generated text.
We evaluate using \mbox{Bleu-1}, \mbox{Bleu-4} \cite{papineni2002bleu}, Meteor \cite{denkowski:lavie:meteor-wmt:2011}, and \mbox{Rouge-L} \cite{Rouge}, using two references for each question,\footnote{We lowercase both the candidates and the references and remove the end of sentence marker and the final full stop.} except for the human baseline where we evaluate one reference against the other.
We also evaluate our models using a ranking metric. This allows us to evaluate how good our model is at reading comprehension regardless of how good it is at generating answers. We rank answers for questions associated with the same summary/story and compute the mean reciprocal rank (MRR).%
\footnote{MRR is the mean over examples of $1/r$, where $r\in\{1,2, \ldots \}$ is the rank of the correct answer among candidates.}

\section{Baselines and Oracles}
\label{sec:baselinebenchmarks}

In this section, we show that NarrativeQA presents a challenging problem for current approaches to reading comprehension by evaluating several baselines based on information retrieval (IR) techniques and neural models. Since neural models use quite different processes for generating answers (e.g., predicting a single word or entity, selecting a span of the document context, or open generation of the answer sequence), we present results on each. We also report the human performance by scoring the second reference answer against the first.

\subsection{Simple IR Baselines}
\label{sec:ir_baselines}

We consider basic IR baselines which retrieve an answer by selecting a span of tokens from the context document based on a similarity measure between the candidate span and a query. We compare two queries: the question and (as an oracle) the gold standard answer. The answer oracle provides an upper bound on the performance of span retrieval models, including the neural models discussed below.
When using the question as the query, we obtain generalization results of IR methods. Test set results are computed by extracting either 4-gram, 8-gram, or full-sentence spans according to the best performance on the validation set.\footnote{Note that we do not consider the span's context when computing the MRR for IR baselines, as the candidate spans (i.e. all answers to questions on the story) are given and simply ranked by their similarity to the query.}

We consider three similarity metrics for extracting spans: \mbox{Bleu-1}, Rouge-L, and the cosine similarity between bag-of-words embedding of the query and the candidate span using pre-trained GloVe word embeddings~\cite{glove}.

\subsection{Neural Benchmarks}
\label{sec:neural_baselines}

As a first benchmark we consider a simple bi-directional LSTM sequence to sequence (Seq2Seq) model~\cite{sutskever2014sequence} predicting the answer directly from the query. Importantly, we provide no context information from either summary or story. Such a model might classify the question and predict an answer of similar topic or category.

Previous reading comprehension tasks such as CNN/Daily Mail motivated
models constrained to predicting a single token from the input sequence. The AS Reader \cite{kadlec2016text} considers the entire context and predicts a distribution over unique word types.
We adapt the model for sequence prediction
by using an LSTM sequence decoder and choosing a token from the input at each step of the output sequence.

As a span-prediction model we consider a simplified version of the Bi-Directional Attention Flow network~\cite{seo2016bidirectional}. We omit the character embedding layer and learn a mapping from words to a vector space rather than making use of pre-trained embeddings; and we use a single layer bi-directional LSTM to model interactions among context words conditioned on the query (modelling layer). As proposed, we adopt the output-layer tailored for span-prediction and leave the rest unchanged.
It was not our aim to use the state-of-the-art model for other datasets but rather to provide a strong benchmark.

\newcommand{\TBF}[1]{\textbf{\footnotesize #1}}
\newcommand{\ASRa}{Attention Sum Reader}
\begin{table*}[t!]
\small
\footnotesize
\centering
\begin{tabular}{llcccccccccc@{}}
    \toprule
    \multicolumn{2}{@{}l}{{\bf Model}}  & \multicolumn{5}{c}{{\bf Validation / Test}} \\
                                        & & {\bf Bleu-1}      & {\bf Bleu-4}      & {\bf Meteor}      & {\bf Rouge-L}     & {\bf MRR} \\
    \midrule
    \multicolumn{2}{@{}l}{{\bf IR Baselines}} \\
    & Bleu-1 given question (1 sentence)             & 10.48/10.75 &  3.02/ 3.34 & 11.93/12.33 & 14.34/14.90 & 0.176/0.171 \\
    & Rouge-L given question (8-gram)                & 11.74/11.01 &  2.18/ 1.99 &  7.05/ 6.50 & 12.58/11.74 & 0.168/0.161 \\
    & Cosine given question (1 sentence)             &  7.49/ 7.51 &  1.88/ 1.97 & 10.18/10.35 & 12.01/12.28 & 0.170/0.171 \\
    & Random rank                          &             &             &             &             & 0.133/0.133 \\
    \midrule
    \multicolumn{2}{@{}l}{{\bf Neural Benchmarks}} \\
    & Seq2Seq (no context)                 & 16.10/15.89 &  1.40/ 1.26 &  4.22/ 4.08 & 13.29/13.15 & 0.211/0.202\\
    & \ASRa                             & 23.54/23.20 &  5.90/ 6.39 &  8.02/ 7.77 & 23.28/22.26 & \TBF{0.269/0.259} \\
    & Span Prediction                      & \TBF{33.45/33.72} & \TBF{15.69/15.53} & \TBF{15.68/15.38} & \TBF{36.74/36.30} & --- \\
    \midrule
    \multicolumn{2}{@{}l}{{\bf Oracle IR Models}} \\
    & Bleu-1 given answer (ans. length)          & 54.60/55.55 & 26.71/27.78 & 31.32/32.08 & 58.90/59.77 & 1.000/1.000 \\
    & Rouge-L given answer (ans. length)   & 52.94/54.14 & 27.18/28.18 & 30.81/31.50 & 59.09/59.92 & 1.000/1.000 \\
    & Cosine given answer (ans. length)    & 46.69/47.95 & 24.25/25.25 & 27.02/27.81 & 44.64/45.66 & 0.836/0.838 \\
    \midrule
    & Human (given summaries)                                & 44.24/44.43 & 18.17/19.65 & 23.87/24.14 & 57.17/57.02 & ---   \\
    \bottomrule
\end{tabular}
\caption{Experiments on summaries. Higher is better for all metrics.
Sections~\ref{sec:ir_baselines} and \ref{sec:neural_baselines} explain the IR and neural models, respectively.}
\label{tab:summary_exp}
\end{table*}

Span prediction models can be trained by obtaining supervision on the training set from the oracle IR model. We use start and end indices of the span achieving the highest Rouge-L score with respect to the reference answers as labels on the training set. The model is then trained to predict these spans by maximizing the probability of the indices.

\subsection{Neural Benchmarks on Stories}

The design of the NarrativeQA dataset makes the straight-forward application of the existing neural architectures computationally infeasible, as this would require running an recurrent neural network on sequences of hundreds of thousands of time steps or computing a distribution over the entire input for attention, as is common.

We split the task into two steps: first, we retrieve a small number of relevant passages from the story using an IR system, and subsequently, apply one of the neural models above on the resulting document. The question becomes the query for retrieval.
This IR problem is much harder that traditional document retrieval, as the documents, the passages here, are very similar, and the question is short and entities mentioned likely occur many times in the story.

Our retrieval system considers chunks of 200 words from story and computes representations for all chunks and the query. We then select a varying number of such chunks based on their similarity to the query. We experiment with different representations and similarity measures in Section \ref{sec:experiments}. Finally, we concatenate the selected chunks in the correct temporal order and insert delimiters between them to obtain a much shorter document. For span prediction models, we then further select a span from the retrieved chunks as described in Section~\ref{sec:neural_baselines}.

\begin{table*}
\centering
\small
\footnotesize
\begin{tabular}{lp{5.5cm}cccccccccc}
    \toprule
    \multicolumn{2}{@{}l}{{\bf Model}}  & \multicolumn{5}{c}{{\bf Validation / Test}} \\
                                        & & {\bf Bleu-1}      & {\bf Bleu-4}      & {\bf Meteor}      & {\bf Rouge-L}     & {\bf MRR} \\    
    \midrule
    \multicolumn{2}{@{}l}{{\bf IR Baselines}} \\
    & Bleu-1 given question (8-gram)                 &  6.73/ 6.52 &  0.30/ 0.34 &  3.58/ 3.35 &  6.73/ 6.45 & 0.176/0.171 \\
    & Rouge-L given question (1 sentence)            &  5.78/ 5.69 &  0.25/ 0.32 &  3.71/ 3.64 &  6.36/ 6.26 & 0.168/0.161 \\
    & Cosine given question (8-gram)                 &  6.40/ 6.33 &  0.28/ 0.29 &  3.54/ 3.28 &  6.50/ 6.43 & 0.171/0.171 \\
    & Random rank                          &             &             &             &             & 0.133/0.133 \\
    \midrule
    \multicolumn{2}{@{}l}{{\bf Neural Benchmarks}} \\
    & \ASRa\ given 1 chunk & 16.95/16.08 & 1.26/1.08 & 3.84/3.56 & 12.12/11.94 & 0.164/0.161 \\
    & \ASRa\ given 2 chunks & 18.54/17.76 & 0.0/1.1 & 4.2/4.01 & 13.5/12.83 & 0.169/0.169 \\
    & \ASRa\ given 5 chunks & 18.91/18.36 & 1.37/1.64 & 4.48/4.24 & 14.47/13.4 & 0.171/0.173 \\
    & \ASRa\ given 10 chunks & \TBF{20.0/19.09} & 2.23/1.81 & 4.45/4.29 & \TBF{14.47/14.03} & 0.182/0.177 \\
    & \ASRa\ given 20 chunks & 19.79/19.06 & \TBF{1.79/2.11} & \TBF{4.6/4.37} & 14.86/14.02 & \TBF{0.182/0.179} \\
    & Span Prediction & 5.82/5.68 & 0.22/0.25 & 3.84/3.72 & 6.33/6.22 & --- \\
    \midrule
    \multicolumn{2}{@{}l}{{\bf Oracle IR Models}} \\
    & Bleu-1 given answer (ans. length)    & 41.81/42.37 &  7.03/ 7.70 & 19.10/19.52 & 46.40/47.15 & 1.000/1.000 \\
    & Rouge-L given answer (ans. length)   & 39.17/39.50 &  7.81/ 8.46 & 18.13/18.55 & 48.91/49.94 & 1.000/1.000 \\
    & Cosine given answer (4-gram)         & 38.21/38.92 &  7.78/ 8.43 & 12.58/12.60 & 31.24/31.70 & 0.842/0.845 \\
    \midrule
    & Human (given summaries)                & 44.24/44.43 & 18.17/19.65 & 23.87/24.14 & 57.17/57.02 & ---   \\
    \bottomrule
\end{tabular}
\caption{Experiments on full stories. Each chunk contains 200 tokens. Higher is better for all metrics.
Sections~\ref{sec:ir_baselines} and~\ref{sec:neural_baselines} explain the IR and neural models, respectively. Note that the human scores are based on answering questions given summaries, same as in Table~\ref{tab:summary_exp}.
}
\label{tab:story_exp}
\end{table*}

\section{Experiments}
\label{sec:experiments}
In this section, we describe the data prepraration methodology we used, and experimental results on the summary-reading task as well as the full story task.

\subsection{Data Preparation}
The provided narratives contain a large number of named entities (such as names of characters or places). Inspired by~\newcite{nips15hermann}, we replace such entities with markers, such as \texttt{@entity42}. These markers are permuted during training and testing so that none of their embeddings learn a specific entity's representation. This allows us to build representations for entities from stories that were never seen in training, since they are given a specific identifier (to differentiate them from other entities in the document) from a set of generic identifiers re-used across documents. Entities are replaced according to a simple heuristic based on capital first character and the respective word not appearing in lowercase. 

\subsection{Reading Summaries Only}

Reading comprehension of summaries is similar to a number of previous reading comprehension tasks where questions were constructed based on the context document.
However, plot summaries tend to contain more intricate event time lines and a larger number of characters, and in this sense, are more complex to follow than news articles or paragraphs from Wikipedia.
See Table~\ref{tab:summary_exp} for the results.

Given that questions were constructed based on the summaries, we expected that both neural models and span-selection models would perform well. This is indeed the case, with the neural span prediction model significantly outperforming all other proposed methods. However, there remains a significant room for improvement when compared with the oracle and human scores.

Both the plain sequence to sequence model and the AS Reader, successfully applied to the CNN/DailyMail reading comprehension task, also perform well on this task. We observe that the AS Reader tends to copy subsequent tokens from the context, thus behaving like a span prediction model.
An additional inductive bias results in higher performance for the span prediction model. Similar observations between AS Reader and span models have also been made
by~\newcite{wang2016machine}.

Note that we have tuned each model separately on the development set twice, once selecting the best model based on Rouge-L and report the first four metrics, and a second time selecting based on MRR.

\subsection{Reading Full Stories Only}
\label{sec:doc_results}

Table~\ref{tab:story_exp} summarizes the results on the full NarrativeQA task, where the context documents are full stories.
As expected (and desired), we observe a decline in performance of the span-selection oracle IR model, compared with the results on summaries. This is unsurprising as the questions 
were constructed on summaries and confirms the initial motivation for designing this task.
As previously, we considered all spans of a given length across the entire story for this model.
For short answers of one or two words---typically main characters in a story---the candidate, i.e.\ the closest span to the reference answer, is easily found due to being mentioned throughout the text.
For longer answers it becomes much less likely, compared to the summaries, that a high-scoring span can be found in the story. Note that this distinguishes NarrativeQA from many of the reviewed datasets.

In our IR plus neural two-step approach to the task, we first retrieve relevant chunks of the stories and then apply existing reading comprehension models. We use the questions to guide the IR system for chunk extraction, with the results of the standalone IR baselines giving an indication of the difficulty of this aspect of the task.
The retrieval quality has a direct effect on the performance of all neural models; a challenge which models on summaries are not presented with.
We considered several approaches to chunk selection: 
we retrieve chunks based on the highest Rouge-L or Bleu-1 scoring span with respect to the question in the story; comparing topic distributions from an LDA model \cite{blei2003latent} between questions and chunks according to their symmetric Kullback--Leibler divergence. Finally, we also consider the cosine similarity of TF-IDF representations. We found that this approach lead to the best performance of the subsequently applied model on the validation set, irrespective of the number of chunks.
Note that we used the answer as the query on the training, and the question for validation and test.

Given the retrieved chunks, we experimented with several neural models using them as context. The AS Reader, which was the better-performing model on the summaries task, underperforms the simple \mbox{no-context} Seq2Seq baseline (shown in Table~\ref{tab:summary_exp}) in terms of MRR. While is does slightly better on the other metrics, it clearly fails to make use of the retrieved context to gain a distinctive margin over the no-context Seq2Seq model. Increasing the number of retrieved chunks, and thereby recall of possibly relevant parts of the story, had only a minor positive effect.
The span prediction model---which here also uses selected chunks for context---does especially poorly in this setup. While this model provided the best neural results on the summaries task, we suspect that its performance was particularly badly hurt by the fact that there is so little lexical and grammatical overlap between the source of the questions (summaries) and the context provided (stories). As with the AS Reader, we observed no significant differences for varying number of chunks.

These results leave us a large gap to human performance, highlighting the success of our design objective to build a task that is realistic and straight-forward for humans while very difficult for current reading comprehension models.

\section{Qualitative Analysis and Challenges}
\label{sec:analysis}

We find that the proposed dataset meets the desiderata we set out in Section~\ref{sec:desiderata}. In particular, we constructed a dataset with a number of long documents, characterised by good lexical coverage and diversity. The questions and answers are human generated and natural sounding. And, based on a small manual examination (of `Ghostbusters~II', `Airplane', `Jacob's Ladder'), only a small number of questions and answers are shallow paraphrases of sentences in the full document. Most questions require reading segments at least several paragraphs long, and in some cases even multiple segments spread throughout the story.

\armageddonexample

Computational challenges identified in Section~\ref{sec:doc_results} naturally suggest a retrieval procedure as the first step.
We found that the retrieval is challenging even for humans not familiar with the presented narrative. In particular, the task often requires referring to larger parts of the story, in addition to knowing at least some background about entities. This makes the search procedure, based on only a short question, a challenging and interesting task in itself. 

We show example question--answer pairs
in Figures \ref{fig:example1}, \ref{fig:example2}, \ref{fig:example3}. These examples were chosen from a small set of manually annotated question--answer pairs to be representative of this collection. In particular, the examples show that larger parts of the story are required to answer questions. Consider Figure~\ref{fig:example3}. While the relevant paragraph depicting the injury appears early on, it is not until the next snippet (which appears at the end of the narrative) that the lethal consequences of the injury are revealed. This illustrates an iterative reasoning process as well as extremely long temporal dependencies we encountered during manual annotation. As shown in Figure~\ref{fig:example1}, reading comprehension on movie scripts requires understanding of written dialogue. This is a challenge as dialogue is typically non-descriptive, whereas the questions were asked based on descriptive summaries, requiring models to ``read between the lines''.

We expect that understanding narratives as complex as those presented in NarrativeQA
will require transferring text understanding capability from other supervised learning tasks.

\section{Related Work}

This paper is the first large-scale question answering dataset on full-length books and movie scripts. However, although we are the first to look at the QA task, learning to understand books through other modeling objectives has become an important subproblem in NLP. These include high level plot understanding through clustering of novels \cite{D17-1200} or summarization of movie scripts
\cite{gorinski-lapata:2015:NAACL-HLT}, to more fine grained processing by inducing character types \cite{bamman-underwood-smith:2014:P14-1,bamman2014learning},
understanding relationships between characters \cite{N16-1180,%
iyyer2017unsupcharact}%
, or understanding plans, goals, and narrative structure in terms of abstract narratives
\cite{schank+abelson77,%
wilensky1978jmm,DBLP:journals/cogsci/BlackW79,Chambers:2009:ULN:1690219.1690231}.
In computer vision, the MovieQA dataset \cite{MovieQA} fulfills a similar role as NarrativeQA. It seeks to test the ability of models to comprehend movies via question answering, and part of the dataset includes full length scripts.

\section{Conclusion}

We have introduced a new dataset and a set of tasks for training and evaluating reading comprehension systems, born from an analysis of the limitations of existing datasets and tasks. While our QA task resembles tasks provided by existing datasets, it exposes new challenges because of its domain: fiction. Fictional stories---in contrast to news stories---are self-contained and describe richer set of entities, events, and the relations between them. We have a range of tasks, from simple (which requires models to read \emph{summaries} of books and movie scripts, and generate or rank fluent English answers to human-generated questions) to more complex (which requires models to read the full \emph{stories} to answer the questions, with no access to the summaries).

\jacobexample

In addition to the issue of scaling neural models to large documents, the larger tasks are significantly more difficult as questions formulated based on one or two sentences of a summary might require appealing to possibly discontiguous sentences or paragraphs from the source text. This requires potential solutions to these tasks to jointly model the process of searching for information (possibly in several steps) to serve as support for generating an answer, alongside the process of generating the answer entailed by said support. End-to-end mechanisms for both searching for information, such as attention, do not scale beyond selecting words or $n$-grams in short contexts such as sentences and small documents. Likewise, neural models for mapping documents to answers, or determining entailment between supporting evidence and a hypothesis, typically operate on the scale of sentences rather than sets of paragraphs. 

We have provided baseline and benchmark results for both sets of tasks, demonstrating that while existing models give sensible results out of the box on summaries, they do not get any traction on the book-scale tasks. Having given a quantitative and qualitative analysis of the difficulty of the more complex tasks, we suggest research directions that may help bridge the gap between existing models and human performance. Our hope is that this dataset will serve not only as a challenge for the machine reading community, but as a driver for the development of a new class of neural models which will take a significant step beyond the level of complexity which existing datasets and tasks permit.

\bibliography{tacl}
\bibliographystyle{acl2012}

\end{document}